# Offline Reconstruction of Missing Vehicle Trajectory Data from 3D LIDAR

Cem Sazara, Reza Vatani Nezafat, Mecit Cetin

*Abstract*— **LIDAR has become an important part of many autonomous vehicles with its advantages on distance measurement and obstacle detection. LIDAR produces point clouds which have important information about surrounding environment. In this paper, we collected trajectory data on a two lane urban road using a Velodyne VLP-16 Lidar. Due to dynamic nature of data collection and limited range of the sensor, some of these trajectories have missing points or gaps. In this paper, we propose a novel method for recovery of missing vehicle trajectory data points using microscopic traffic flow models. While short gaps (less than 5 seconds) can be recovered with simple linear regression, and longer gaps are recovered with the proposed method that makes use of car following models calibrated by assigning weights to known points based on proximity to the gaps. Newell's, Pipes, IDM and Gipps' car following models are calibrated and tested with the ground truth trajectory data from LIDAR and NGSIM I-80 dataset. Gipps' calibrated model yielded the best result.**

## I. INTRODUCTION

Environment detection is one of the essential elements of autonomous vehicles. Environment can be composed of moving cars, parked cars, pedestrians, trees, etc. Autonomous vehicles need sensors and camera systems to navigate while avoiding obstacles. LIDARs have become a popular sensor for autonomous vehicles. LIDAR is a remote sensing technology that measures surrounding area with lasers and provides rich data about the spatial information around the vehicle. Laser beams are fired at surfaces and returned beams are collected and measured. In this paper, we are using a Velodyne 3D VLP-16 LIDAR sensor. The sensor has 16 laser beams with different pitch angles placed as a rotating unit. It supports 360 degree horizontal field of view and 30 degree vertical field of view, and it produces approximately 30,000 points in a single scan with a range of 100m. Rotation frequency can be adjusted between 5-20Hz [1]. In this paper, it is selected as 10Hz. Therefore, in every 0.1 seconds we are able to see a snapshot of the surrounding environment. Each snapshot can also be called a scan. LIDAR data are collected in urban area and trajectories for the vehicles in the front of the probe or ego vehicle are constructed.

Most of the data for traffic flow theory has been collected with fixed point detectors such as loop detectors and fixed cameras that lack accuracy to capture microscopic relationships between vehicles. Fixed point detectors can only collect data from certain locations which reduces the diversity of the data. GPS enabled phones, smart phones and equipped vehicles emerged as an alternative to fixed point collectors. Data from these sources are used to construct leader and follower vehicle trajectories [2], [3], [4]. Another way of mobile data collection is with LIDARs. Collecting data with high resolution LIDARs allows scientists to capture a wide variety of vehicle interactions under different driving conditions. Given the abundance of data, LIDAR data processing is a challenging task and there are many different methods in the literature for object classification and tracking [5-8]. Here, a simpler LIDAR processing method is applied since the main focus is studying microscopic relations between vehicles. Data are collected with a vehicle equipped with a 3D LIDAR, GPS, accelerometer and a dash camera. In addition to the data abundance problem, in contrary, missing data is another problem that arises mostly for distant obstacles due to dynamic moving nature of data collection and sensor range limitations. We are interested in missing data problem. So far, LIDAR data has been mostly studied for obstacle detection and tracking purposes but there is also rich information useful for traffic flow theory studies [9-11]. Given this motivation, we propose a novel method that recovers missing data points in vehicle trajectories. Complete trajectories are important source for traffic flow studies. Our method exploits the advantage of car following models in reconstruction of missing data in vehicle trajectories as these models specifically explain car following phenomena. Therefore, missing data are recovered using the interaction between leader and follower vehicles. For this purpose, different car following models are calibrated and missing data points are estimated. The method is tested with the ground truth LIDAR trajectories and NGSIM I-80 dataset [12].

## II. RELATED WORK

Our trajectory recovery method is closely related to car following models. These microscopic traffic models describe longitudinal motion of vehicles such as accelerating and decelerating in response to traffic situation and keeping a safe distance with other vehicles. Microscopic models assume that drivers' reactions are heavily dependent on leading vehicles and they explain this phenomena in terms of leader-follower vehicle pairs and many of the models include numerous parameters. Calibration of car following models consists of selecting optimum model parameters that minimize a given cost or error function. Driving behavior is an important concept and roadway design, capacity, travel time and road safety are directly influenced by human driving behavior. Inter-driver and Intra-driver heterogeneity concepts emerged due to human driving behavior. Inter-driver heterogeneity occurs due to different driving styles between drivers. Ossen et al [13] studied multiple car following models and concluded that performance of more complex models differs between drivers. In a later study [14], Ossen et al. studied trajectories of cars and trucks and concluded that reactions of the drivers of the cars showed more fluctuations than the drivers of the trucks. Duret et al. [15] proposed an estimation method for inter-vehicle heterogeneity and calibrated Newel's car following model. Intra-driver heterogeneity is another important concept and it occurs because drivers can react to different situations differently at different times. Therefore, drivers' reactions may change depending on the traffic conditions and time. In [16], authors conclude that a larger part of calibration errors comes from intra-driver heterogeneity rather than inter-driver heterogeneity. Wagner [17] analyzed fluctuations in car-following and concluded that headway fluctuations occur due to internal stochasticity of the driver. Wang et al [18] confirmed that intra-driver heterogeneity occurred between acceleration and deceleration parts of the car following process. Given these important results, we calibrated individual car following models for each gap in our data using the known information before and after the gap. Known points around the gap are weighted based on the distance from start or end of the gap depending on the side of the gap. Therefore, we are able to capture intra-driver behavior and make better estimation of the gap points. Estimated points start to be used from the beginning of the gap. Towards the end of the gap, we introduced some rules so that estimated points do not deviate too much from the end of the gap.

### III. DATA

#### A. LIDAR Data

Our data acquisition platform consists of LIDAR (VLP-16), GPS and dash camera. Velodyne LIDAR is mounted on the roof the car and data are collected on urban roads with multiple trips. Our LIDAR provides rich 3D data points for every 0.1 seconds which we will call a scan. An example top-view LIDAR scan image is given in Fig. 1. Right-handed coordinate system is used to represent 3D points where +y indicates the direction of vehicle's forward movement. Every scan is filtered based on elevation thresholds, lane thresholds and direction. Elevation thresholds help remove ground points and points that belong to high structures such as trees, wires etc. Lane threshold makes sure that we only consider vehicles that are on the same lane as the data collection vehicle. Because of moving in only forward direction, points in front of the vehicle are considered. As a future work, multi-lane processing and tracking [6-8] will be implemented. Points satisfying these conditions are collected and clustered based on distance proximity. Minimum 2D distance from the cluster to data collection vehicle is recorded. Using the synchronized accelerometer and distance information, trajectories are constructed. Recorded vehicle is considered as the leader vehicle and our vehicle is the follower vehicle for each trajectory pair. Some manual work is applied by looking at dash cam records to verify that lane change did not occur and the same car was followed for each trajectory. When we analyzed the trajectories, we realized that some of the trajectories have gaps due to range limitations of the sensor. For example, leader vehicle goes out of the range of LIDAR and enters back again. Our proposed method provides a solution to this problem and recovers missing data in trajectories. The method is explained

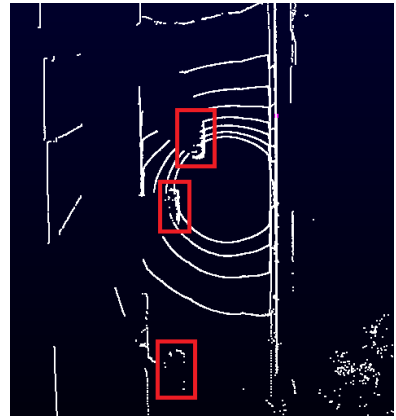

Figure 1. Top-view scan with vehicles shown in red bounding-boxes.

in detail in section IV.

#### B. NGSIM Data

In addition to LIDAR data, we used NGSIM I-80 dataset [12] to test our trajectory gap recovery method. NGSIM dataset was collected on eastbound Interstate 80 in Emeryville, CA in 2005 and has been a great resource for researches in developing and validating algorithms and models. The investigated area is approximately 500 meter (1,640 feet) long and consists of 6 lanes. Data are in total 45 minutes long and divided into three 15-minute sections. Data in this paper is taken from the first 15-minute part of the total trajectory data. We applied the following rules in extracting leader-follower trajectories from this dataset.

- Length of the trajectory should be long enough to reflect intra-driver behavior. Minimum length is selected as 50 seconds.
- Lane change should be filtered out. Studied car following models assume no lane change.
- Leftmost (HOV) and rightmost (ramp) lanes are filtered out.

### IV. APPROACH

After LIDAR trajectory data are collected, complete LIDAR trajectories are used along with the NGSIM data to test our method. Since the used LIDAR and NGSIM data don't contain missing points, random gaps are applied on the trajectories. The proposed method fills missing data points in the trajectory data. We assume driving behavior stays the same for at most 5 seconds. Therefore, when the length of the gap is less than 5 seconds, we used linear interpolation. For gaps longer than 5 seconds, we need to use a more complicated method to incorporate intra-driver driver behavior change. For this purpose, we introduced a method that uses car following models with weights applied on points before and after the gap. For each gap, an optimum car following model is found and the trajectory corresponding to the gap is constructed. In order to match the end of the constructed trajectory with the end of the gap, we proposed an algorithm which we call "Smooth transition algorithm". Details of the approach applied on gaps longer than 5 seconds are given below.

*A. Car following models*

Gipps', Intelligent Driving Model, Pipe's and Newell's [23] car following models are implemented for each gap in the data. Equations and calibrated parameters are explained below.

Gipp's model assumes a "safe speed" so that the following vehicle avoids collision even when the leader vehicle suddenly stops and keeps a "safe distance" with the leader vehicle. It also assumes a certain reaction time for the follower vehicle. Its formula is given by

$$v(t+\Delta t) = \min(v + a\Delta t, v_0, v_{safe}(s, v_l)) \quad (1)$$

$$v = \min(v_0, v_{safe}) \quad (2)$$

$$v = \min(v_0, -b\,\Delta t + \sqrt{b^2 \Delta t^2 + v^2 + 2b(s-s_0)}) \quad (3)$$

with parameters speed v, desired speed $v_0$, safe speed $v_{safe}$, deceleration b, reaction time $\Delta t$, distance s and minimum distance $s_0$.

The Intelligent Driver Model (IDM) assumes a "safe distance" and uses a smooth acceleration and deceleration policy. Transition between states are also smooth [20]. It is given by

$$\dot{v} = a\left[1 - \left(\frac{v}{v_0}\right)^\delta - \left(\frac{s^*(v,\Delta v)}{s}\right)^2\right] \quad (4)$$

with the parameters speed v, speed change $\Delta v$, desired speed v0, acceleration exponent δ, acceleration a and desired distance s* and current distance s. Desired distance

$$s^*(v,\Delta v) = s_0 + \max(0, vT + \frac{v\Delta v}{2\sqrt{ab}}) \quad (5)$$

is given as minimum distance $s_0$ and addition of a dynamical term where T is time gap, a is acceleration and b is deceleration. This term allows intelligent following distance.

Pipe's Model [21] assumes a "safe distance" between the leader and follower vehicles which is at least the length of a car for every ten miles per hour. Its equation is given by

$$x(n) = x(n+1) + b + L(n) + s_{min}(n+1) \quad (6)$$

where x(n) and x(n+1) are the positions of the leader and follower vehicles respectively, b is the distance between vehicles when they are at standstill and L(n) is the length of the leader vehicle and $s_{min}$(n+1) is the minimum distance between the vehicles given by

$$s_{min}(n+1) = T.V(n+1) \quad (7)$$

where v(n + 1) is the speed of vehicle n + 1, in m/s and T is the time it takes to travel this minimum distance, in second.

Newell [22] proposed a car following model that makes use of the linear translation of trajectories in time and space. Newell's model assumes that under congested conditions a vehicle will maintain a minimum gap between itself and the car in the front.

$$x_n(t+\tau_n) = x_{n-1}(t) - d_n \quad (8)$$

where $\tau_n$ and $d_n$ are time and distance translations, respectively.

*B. Calibration methodology*

Optimum parameters for the car following models are found by solving an optimization problem with genetic algorithm. Genetic algorithm has been widely used in the literature for the calibration problem and it helps solve both constraint and unconstrained optimization problem given a cost function. Our cost function is sum of absolute error values between car following model headway values and known trajectory points headway values multiplied by individual weights. We applied tri-cube weight function on the points around the gap so that points closer to the gap have more effect on the cost function. Cost function is given by

$$C = \sum_{i=1}^{N} w_i abs(s_{i,model} - s_{i,trajectory}) \quad (9)$$

where N is the total number of points around the gap, $s_{i,model}$ is car following model headway at point i and $s_{i,trajectory}$ is headway of known data point i and $w_i$ is weight for point i. The weight for each point i is calculated by

$$w_i(d,L) = \begin{cases} (1-(d/L)^3)^3 & for\,(d/L) < 1 \\ 0 & for\,(d/L) \geq 1 \end{cases} \quad (10)$$

where d is the distance from the edge of the gap and L is the length of before and after gap areas. Fig. 2 shows the gap and the before and after gap areas and Fig. 3 shows the result of the calibration. Leader data at the before and after gap areas are used to calibrate car following models. In cost function calculations, the weight function in Eq. 10 is used which results in 1 ("Max weight" in Fig. 2.) at the nearest point at the gap edges and yields 0 ("Min weight" in Fig. 2.) at the farthest points of before and after gap areas. The points between the max and min weight points are assigned weights based on the tri-cube function given in Eq. 10. Therefore, closer points have more effect on finding the optimum car following parameters. Fig 3 shows a calibrated gap. We can see that at the end of the gap, we have unconnected end points. This is expected as the car following model is applied starting from the beginning of the gap. We developed an algorithm that reshapes the end part and connects unconnected end points. We call this algorithm "Smooth transition algorithm".

*C. Smooth Transition Algorithm*

We propose a novel algorithm for reshaping and connecting end points of vehicle trajectory gaps. Steps of the algorithm are:

1. Straight lines from the ending edge of the gap are drawn to each estimated point starting from the closer points. Lines are drawn until the difference between the slope of the drawn line and the slope at the edge of the gap is less than a threshold. This makes sure a smooth transition so that speed (slope) does not change drastically. When the slope threshold condition is satisfied, we stop drawing lines. Because we find the start point of the reshape operation. End point is the last point of the estimated points.

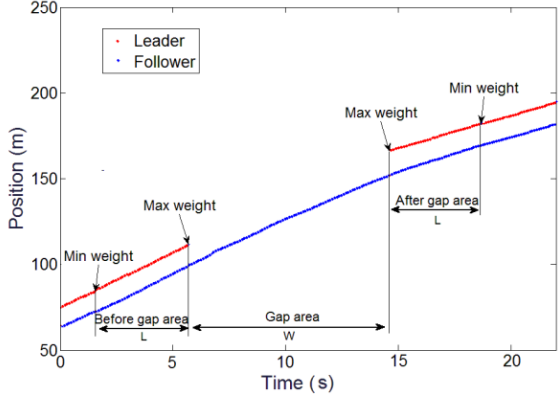

Figure 2. Trajectory gap and before-after areas

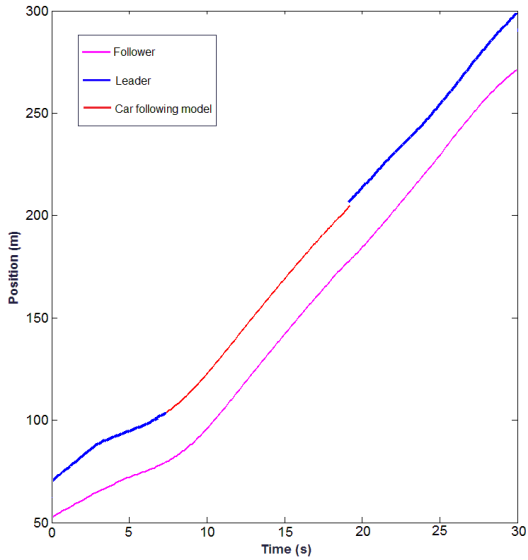

Figure 3. Calibrated gap with unconnected end points

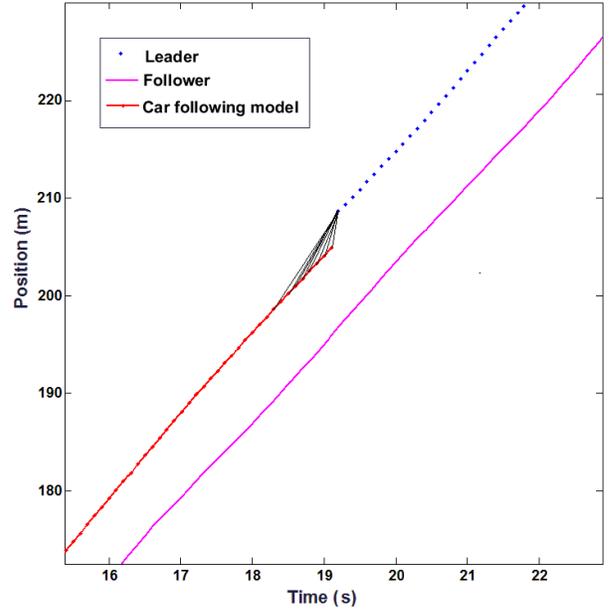

Figure 4. Step 2 - Straight lines from the end of the gap

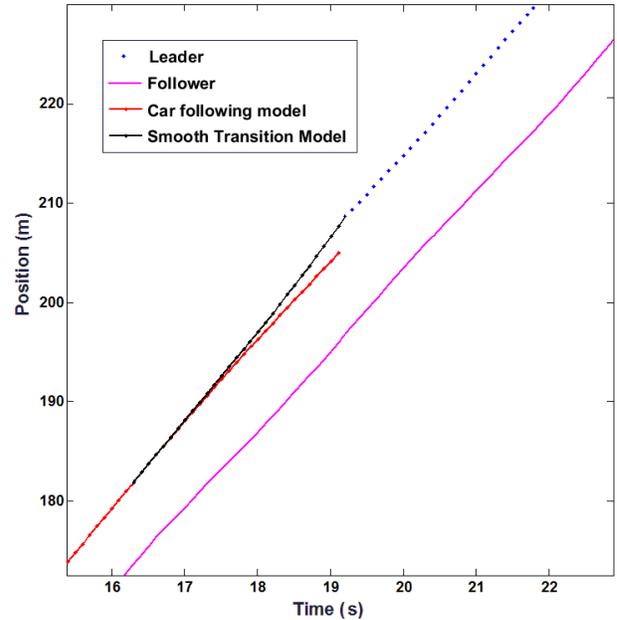

Figure 5. Result of the smooth transition algorithm

2. Reshaping operation is applied. This operation uses both the car following model values and the last drawn line values. We apply a varying ratio between the two values to make sure that when we are away from the end point, car following model value is weighted more. When getting close to the end of the gap, line value is weighted more. At the end, it is weighted 100% end point and we are able to connect the end points. Weighting operation for point n can be given in Eq. 11.

$$y_{stm}[n] = w_n \cdot y_{cfm}[n] + (1 - w_n) \cdot y_{line}[n] \quad (11)$$

where $y_{stm}[n]$ is the smooth transition model value, $y_{cfm}[n]$ is car following model value, $y_{line}[n]$ is line value and $w_n$ is the weight at point n.

Fig. 4 shows step 1 of the algorithm and Fig. 5 shows the result after the algorithm is applied. We can see that smooth transition algorithm is able to fix the unconnected end point problem. For consistency and easy understanding, the areas shown in Fig. 3, Fig. 4 and Fig. 5 are the same.

## V. RESULTS

The proposed method is applied on ground truth LIDAR and NGSIM trajectory data. In order to test the method, we created random gaps of length between 5 and 15 seconds in the complete LIDAR and NGSIM trajectory datasets. We used the data points 5 seconds before and after the gaps in order to capture intra-driver behavior. Gipps', Intelligent Driving Model, Pipe's and Newell's models are calibrated using genetic algorithm with the cost function of Eq. 9. Genetic algorithm is applied with roulette wheel selection, crossover rate 0.7, mutation rate 0.1, population 20 and 50 number of iterations. 112 gaps in NGSIM and 24 gaps in LIDAR data are investigated. Root mean square error (RMSE) and mean absolute percent error (MAPE) are calculated for each gap and car following model for NGSIM and LIDAR data. Fig. 6 and Table 1 summarize calibration errors for the LIDAR and NGSIM data.

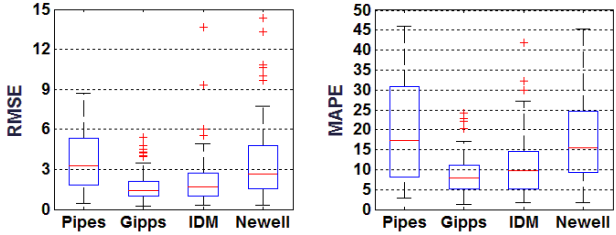

Figure 6. MAPE and RMSE (meters) for gaps LIDAR and NGSIM data

TABLE I. ERROR STATISTICS FOR LIDAR AND NGSIM DATA

| | Car following models | | | | | | | |
|---|---|---|---|---|---|---|---|---|
| | *Pipes* | | *Gipps* | | *IDM* | | *Newell* | |
| | % MAPE | RMSE | % MAPE | RMSE | % MAPE | RMSE | % MAPE | RMSE |
| Min | 2.87 | 0.48 | 1.35 | 0.22 | 1.78 | 0.34 | 1.63 | 0.29 |
| Max | 46.1 | 8.71 | 24.13 | 5.41 | 41.78 | 13.67 | 45.29 | 14.36 |
| Average | 20.27 | 3.63 | 8.95 | 1.75 | 11.21 | 2.26 | 17.79 | 3.65 |
| Median | 17.4 | 3.29 | 7.82 | 1.42 | 9.66 | 1.68 | 15.57 | 2.66 |
| Standard diviation | 13.37 | 2.12 | 5.36 | 1.15 | 7.58 | 2.04 | 10.51 | 2.99 |

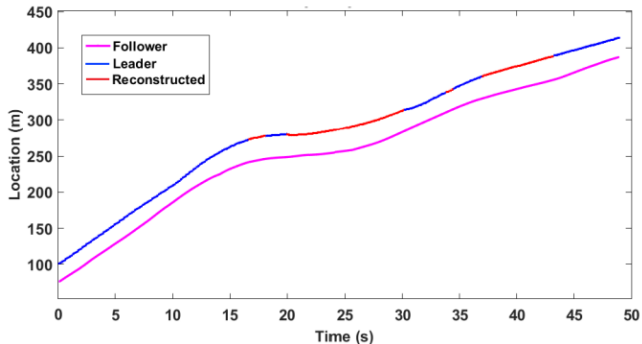

Figure 7. Two gaps on LIDAR data

From Fig. 6, we can see that Gipps and IDM lead to more accurate prediction than others. Also, based on Table 1, Gipps model has a lower standard deviation than IDM and it results in more stable predictions. Therefore, we recommend Gipps model to be used for reconstruction of gaps in trajectories.

In Fig. 7, the proposed method is applied by using Gipps model and two large gaps in the collected LIDAR trajectory data are recovered. Small gaps (<5seconds) are also recovered by linear interpolation.

## VI. CONCLUSION

In this paper, we proposed a novel method to recover missing data in vehicle trajectories. We tested our method with ground truth LIDAR and NGSIM data. The method gave similar results for both of the datasets and Gipps car following model yielded the least error. As a future work, we are going to investigate macroscopic characteristics of traffic flow based on LIDAR data and study driving behavior more explicitly.